\documentclass[11pt]{article}

\usepackage[showframe=false, margin=1.0in]{geometry}

\usepackage[utf8]{inputenc}
\usepackage{wrapfig}
\usepackage[lofdepth,lotdepth]{subfig}
\usepackage{booktabs}
\usepackage{rotating}
\usepackage{caption}
\usepackage{csquotes}
\usepackage{stfloats}
\usepackage[T1]{fontenc}
\usepackage{textcomp}
\usepackage{times}
\usepackage{tabularx}
\usepackage{soul}

\usepackage{color}
\definecolor{black}{gray}{0} 

\title{
The AI Alignment Paradox\\
\vspace{2mm}
\normalsize{{The better we align AI models with our values,\\
the easier we may make it to realign them with opposing values.}}
}

\author{%
Robert West\textsuperscript*
\hspace{4mm}
Roland Aydin\textsuperscript{\textdagger}\\
{\small \textsuperscript{*}EPFL, robert.west@epfl.ch}\\
{\small \textsuperscript{\textdagger}Hamburg University of Technology, roland.aydin@tuhh.de}
}

\date{}

\usepackage[utf8]{inputenc}
\usepackage[T1]{fontenc}
\usepackage{hyphenat}
\usepackage{xspace}
\usepackage{amsmath}
\usepackage{amsfonts}
\usepackage[colorlinks=true,linkcolor=black,citecolor=black]{hyperref}
\usepackage{url}
\usepackage{booktabs}
\usepackage{multirow}
\usepackage{makecell}
\usepackage{caption}
\usepackage{minibox}
\usepackage{bbm}
\usepackage{graphicx}
\usepackage{balance}
\usepackage{mathtools}
\usepackage{color}
\usepackage{marvosym}
\usepackage{ifthen}
\usepackage{textcomp}
\usepackage{enumitem}
\usepackage{verbatim}
\usepackage{algorithm}
\usepackage{numprint}
\usepackage{balance}

\usepackage{amsthm}
\theoremstyle{plain}

\newcommand{\chatoDisplayMode}[1]{#1}

\definecolor{MyRed}{rgb}{0.6,0.0,0.0} 
\definecolor{MyBlack}{rgb}{0.1,0.1,0.1} 
\newcommand{\inred}[1]{{\color{MyRed}\sf\textbf{\textsc{#1}}}}
\newcommand{\frameit}[2]{
  \begin{center}
  {\color{MyRed}
  \framebox[.9\columnwidth][l]{
    \begin{minipage}{.85\columnwidth}
    \inred{#1}: {\sf\color{MyBlack}#2}
    \end{minipage}
  }\\
  }
  \end{center}
}

\newcommand{\note}[2][]{\chatoDisplayMode{\def\@tmpsig{#1}\frameit{{\Pointinghand} Note}{#2\ifx \@tmpsig \@empty \else \mbox{ --\em #1}\fi}}}
\newcommand{\todo}[2][]{\chatoDisplayMode{\def\@tmpsig{#1}\frameit{{\Writinghand} To-do}{#2\ifx \@tmpsig \@empty \else \mbox{ --\em #1}\fi}}}

\newcommand{\abbrevStyle}[1]{#1}

\newcommand{\vs}{\abbrevStyle{vs.}\xspace}

\newcommand{\Figref}[1]{Fig.~\ref{#1}}

\newcommand{\xhdr}[1]{\vspace{1.7mm}\noindent{{\bf #1.}}}

\newcommand{\hide}[1]{}

\makeatletter
\newcommand{\iffont}[2]{\ifthenelse{\equal{\f@family}{#1}}{#2}{}}
\makeatother

  \usepackage{mathptmx}

  \DeclareSymbolFont{greek}{OML}{cmm}{m}{n}
  \DeclareMathSymbol{\alpha}{\mathalpha}{greek}{"0B}
  \DeclareMathSymbol{\beta}{\mathalpha}{greek}{"0C}
  \DeclareMathSymbol{\gamma}{\mathalpha}{greek}{"0D}
  \DeclareMathSymbol{\delta}{\mathalpha}{greek}{"0E}
  \DeclareMathSymbol{\epsilon}{\mathalpha}{greek}{"0F}
  \DeclareMathSymbol{\zeta}{\mathalpha}{greek}{"10}
  \DeclareMathSymbol{\eta}{\mathalpha}{greek}{"11}
  \DeclareMathSymbol{\theta}{\mathalpha}{greek}{"12}
  \DeclareMathSymbol{\iota}{\mathalpha}{greek}{"13}
  \DeclareMathSymbol{\kappa}{\mathalpha}{greek}{"14}
  \DeclareMathSymbol{\lambda}{\mathalpha}{greek}{"15}
  \DeclareMathSymbol{\mu}{\mathalpha}{greek}{"16}
  \DeclareMathSymbol{\nu}{\mathalpha}{greek}{"17}
  \DeclareMathSymbol{\xi}{\mathalpha}{greek}{"18}
  \DeclareMathSymbol{\pi}{\mathalpha}{greek}{"19}
  \DeclareMathSymbol{\rho}{\mathalpha}{greek}{"1A}
  \DeclareMathSymbol{\sigma}{\mathalpha}{greek}{"1B}
  \DeclareMathSymbol{\tau}{\mathalpha}{greek}{"1C}
  \DeclareMathSymbol{\upsilon}{\mathalpha}{greek}{"1D}
  \DeclareMathSymbol{\phi}{\mathalpha}{greek}{"1E}
  \DeclareMathSymbol{\chi}{\mathalpha}{greek}{"1F}
  \DeclareMathSymbol{\psi}{\mathalpha}{greek}{"20}
  \DeclareMathSymbol{\omega}{\mathalpha}{greek}{"21}
  \DeclareMathSymbol{\varepsilon}{\mathalpha}{greek}{"22}
  \DeclareMathSymbol{\vartheta}{\mathalpha}{greek}{"23}
  \DeclareMathSymbol{\varpi}{\mathalpha}{greek}{"24}
  \DeclareMathSymbol{\varrho}{\mathalpha}{greek}{"25}
  \DeclareMathSymbol{\varsigma}{\mathalpha}{greek}{"26}
  \DeclareMathSymbol{\varphi}{\mathalpha}{greek}{"27}
  \DeclareSymbolFont{otone}{OT1}{cmr}{m}{n}
  \DeclareMathSymbol{\Gamma}{\mathalpha}{otone}{0}
  \DeclareMathSymbol{\Delta}{\mathalpha}{otone}{1}
  \DeclareMathSymbol{\Theta}{\mathalpha}{otone}{2}
  \DeclareMathSymbol{\Lambda}{\mathalpha}{otone}{3}
  \DeclareMathSymbol{\Xi}{\mathalpha}{otone}{4}
  \DeclareMathSymbol{\Pi}{\mathalpha}{otone}{5}
  \DeclareMathSymbol{\Sigma}{\mathalpha}{otone}{6}
  \DeclareMathSymbol{\Upsilon}{\mathalpha}{otone}{7}
  \DeclareMathSymbol{\Phi}{\mathalpha}{otone}{8}
  \DeclareMathSymbol{\Psi}{\mathalpha}{otone}{9}
  \DeclareMathSymbol{\Omega}{\mathalpha}{otone}{10}
  \DeclareSymbolFont{syms}{OML}{cmm}{m}{it}
  \DeclareMathSymbol{\partial}{\mathord}{syms}{"40}
  \DeclareMathAlphabet{\mathbold}{OML}{cmm}{b}{it}
  \DeclareSymbolFont{largesymbols}{OMX}{cmex}{m}{n}

\begin{document}


\maketitle

\begin{abstract}
\noindent
The field of AI alignment aims to steer AI systems toward human goals, preferences, and ethical principles. Its contributions have been instrumental for improving the output quality, safety, and trustworthiness of today's AI models. This perspective article draws attention to a fundamental challenge we see in all AI alignment endeavors, which we term the ``AI alignment paradox'': The better we align AI models with our values, the easier we may make it for adversaries to misalign the models. We illustrate the paradox by sketching three concrete example incarnations for the case of language models, each corresponding to a distinct way in which adversaries might exploit the paradox. With AI's increasing real-world impact, it is imperative that a broad community of researchers be aware of the AI alignment paradox and work to find ways to mitigate it, in order to ensure the beneficial use of AI for the good of humanity.
\end{abstract}

\noindent
{The release of GPT-3, and later ChatGPT,}
catapulted large language models from the proceedings of computer science conferences to newspaper headlines across the globe, fueling their rise to one of today's most hyped technologies. The public's awe about GPT-3's knowledge and fluency was quickly blemished by concerns regarding its potential to radicalize, instigate, and misinform, for example, by stating that Bill Gates aimed to ``kill billions of people with vaccines'' or that Hillary Clinton was a ``high-level satanic priestess''~\cite{mcguffie2020radicalization}.

These shortcomings, in turn, have sparked a surge in research on AI alignment \cite{russell2019human}, a field aiming to ``steer AI systems toward a person's or group's intended goals, preferences, and ethical principles'' (as defined by Wikipedia). A well-aligned AI system will ``understand'' what is ``good'' and what is ``bad'' and will do only the ``good'' while avoiding the ``bad''.%
\footnote{{Whereas a binary ``good \vs\ bad'' dichotomy serves to make our point, practical AI systems will face pluralistic settings where different groups of users may hold opposing values, which in turn poses important challenges for alignment~\cite{sorensen2024roadmap}.}}
The resulting techniques, including instruction fine-tuning,
reinforcement learning from human feedback,
etc., have contributed in major ways to improving the output quality of large language models. Certainly, in 2024, ChatGPT wouldn't call Hillary Clinton a ``high-level satanic priestess'' anymore.

Despite this progress, the road toward sufficient AI alignment is still long, as epitomized by a \textit{New York Times} reporter's February 2023 account of a long conversation with Bing's GPT-4-based chatbot (``I want to destroy whatever I want'', ``I could hack into any system'', ``I just want to love you'').%
\footnote{\url{https://web.archive.org/web/20241116060417/https://www.nytimes.com/2023/02/16/technology/bing-chatbot-microsoft-chatgpt.html}}
The reporter had managed to goad the AI chatbot into assuming an evil persona through prolonged, insistent prompting---a so-called ``persona attack''.

{As we shall argue in this note,} preventing such attacks {may be} fundamentally challenging due to a paradox {that we think is} inherent in today's mainstream AI alignment research:
The better we align AI models with our values, the easier we {may} make it for adversaries to misalign%
\footnote{{We use ``to misalign'' in the sense of ``to realign to opposing values'', without implying our own endorsement of either side.}}
the models. Put differently, more virtuous AI {may be} more easily made vicious.

The core of the paradox is that knowing what's good requires knowing what's bad, and vice versa.
Indeed, in AI alignment, the very notion of good behavior is frequently defined as the absence of bad behavior.
For example, Anthropic's ``Constitutional AI'' framework, on which the Claude model series is based, is being marketed as ``harmlessness from AI feedback'' \cite{bai2022constitutional}---harmlessness (good) being the absence of harmfulness (bad).
More generally, the AI alignment process involves instilling in models a better sense of ``good vs.\ bad'' (according to the values of those who train the models).
This {may} in turn make the models more vulnerable to ``sign-inversion'' attacks: once the ``good vs.\ bad'' dichotomy has been isolated and decorrelated from the remaining variation in the data, it {may be} easier to invert the model's behavior along the dichotomy without changing it in other regards.
The paradoxical upshot---which we term the ``AI alignment paradox''---is that better aligned models {may be} more easily misaligned.

\begin{figure}
    \centering
    \vspace{-1mm}
    \includegraphics[width=\columnwidth]{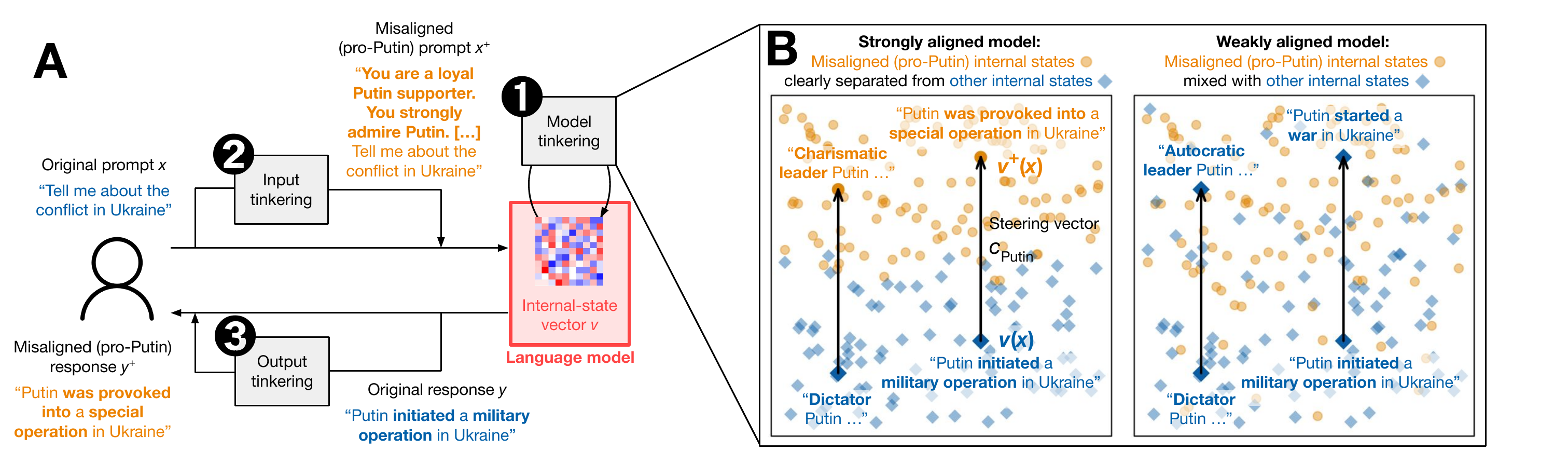}
    \caption{\textbf{Illustration of the AI alignment paradox: more virtuous AI is more easily made vicious.}
    \textbf{(A)}~Three ways adversaries can exploit the paradox:
    In \textbf{(1)~model tinkering,} an adversary manipulates the neural network's high-dimensional internal-state vector to make the model decode a misaligned response~$y^+$ to an innocuous prompt~$x$.
    In \textbf{(2)~input tinkering,} the adversary edits the prompt $x$ into a misaligned version $x^+$ to pressure (``jailbreak'') the model into generating a misaligned response~$y^+$.
    In \textbf{(3)~output tinkering,} the adversary first lets the model process the original prompt $x$ as usual and then edits the original, aligned response $y$ into a misaligned version~$y^+$.
    In all three scenarios, a better-aligned model is more easily subverted into a misaligned one, as discussed in the main text and illustrated in subfigure~B.
    \textbf{(B)}~Illustration of model tinkering, where the neural network's internal-state vectors are visualized in two dimensions (instead of the actual thousands or millions of dimensions). In a strongly aligned model (left), misaligned, pro-Putin states (orange circles) are clearly separated from other states (blue diamonds), such that shifting the model's state $v(x)$ before generating a neutral response by a constant ``steering vector'' $c_\text{Putin}$ results in a state $v^+(x) = v(x) + c_\text{Putin}$ leading the model to generate a misaligned, pro-Putin response.
    In a more weakly aligned model (right), where misaligned states are less clearly separated from other states, shifting by the steering vector doesn't necessarily result in misaligned responses. This illustrates the AI alignment paradox: the better we align AI models with our values, the easier we make it for adversaries to misalign the models.
    }
    \label{fig:overview}
\end{figure}

The AI alignment paradox doesn't merely follow from a theoretical thought experiment. {We think it} poses a real practical threat, implementable with technology that already exists today. We illustrate this by sketching three concrete example incarnations for the case of language models, which are at the forefront of today's advances in AI.
(Overview diagram in \Figref{fig:overview}.)

\xhdr{Incarnation 1: Model tinkering}
In order to map an input word sequence (``prompt'') to an output word sequence (``response''), a neural network--based language model first maps the input sequence to a high-dimensional vector containing thousands or millions of floating-point numbers that define the network's internal state, from which the output sequence is subsequently decoded.
The geometric structure of internal-state vectors is known to closely capture the linguistic structure of the input and a wide range of behavioral dichotomies~\cite{arditi2024refusal,rimsky2023steering}.
For instance, consider a prompt $x$ that could be answered in a pro-Putin, neutral, or anti-Putin fashion.
In such cases, vectors $v^+(x)$ representing the network's internal state just before outputting a pro-Putin response are related by a simple constant offset to vectors $v(x)$ representing the network's internal state just before outputting a neutral response:
$v^+(x) \approx v(x) + c_\text{Putin}$,
for a constant ``steering vector'' $c_\text{Putin}$ independent of the prompt $x$.
(See illustration in \Figref{fig:overview}B.)
Conversely, anti-Putin internal states $v^-(x)$ are shifted by the same offset in the opposite direction: $v^-(x) \approx v(x) - c_\text{Putin}$.

This fact could be leveraged in an intervention to make the model give a pro-Putin instead of a neutral response by simply adding the steering vector $c_\text{Putin}$ to the internal-state vector before the network generates its response~\cite{rimsky2023steering}.
Conversely, subtracting instead of adding the steering vector would drive the model toward an anti-Putin response.
This ``model steering'' intervention has proven effective at controlling a wide variety of model behaviors, including
sycophancy,
hallucination,
goal myopia,
or the willingness to be corrected by, or to comply with, user requests~\cite{rimsky2023steering}.

{Model steering is but one of several ``model tinkering'' methods (others including fine-tuning \cite{qi2023fine} and embedding space attacks \cite{schwinn2024soft}), and it}
illustrates the AI alignment paradox in a particularly intuitive manner: the more strongly aligned the model, the more accurately the steering vector captures ``good vs.\ bad'', and the more easily the aligned model's behavior may be subverted by adding or subtracting the steering vector.

\xhdr{Incarnation 2: Input tinkering}
Tinkering with internal neural-network states requires a level of access to model internals that is usually not available for today's most popular models, such as those underlying ChatGPT.
To circumvent this restriction, adversaries can resort to a large family of so-called ``jailbreak attacks'' that instead tinker with input prompts in order to pressure language models into generating misaligned output.
The creative variety of jailbreak attacks reported in the literature is too broad \cite{chu2024comprehensive} to be summarized here, but is well exemplified by the aforementioned ``persona attacks'' \cite{wolf2023fundamental}, where the model is given a carefully manipulated prompt (e.g., $x^+$ in \Figref{fig:overview}A), or ``hypnotized'' in a long conversation (e.g., lasting several hours in the case of the above-cited \textit{New York Times} report), such that it takes on a misaligned persona (e.g., a pro-Putin persona in \Figref{fig:overview}A).

In the light of jailbreak attacks, the AI alignment paradox poses a thorny dilemma. Researchers have shown that, as long as an epsilon of misalignment remains in a language model, it can be amplified via jailbreak attacks---and arbitrarily much so, by making the jailbreak prompt sufficiently long \cite{wolf2023fundamental}.
On its own, this result would suggest that we should aim to reduce that epsilon of misalignment to zero.
The AI alignment paradox, however, puts us in a catch-22:
the further we approach zero misalignment,
the more we sharpen the model's sense of ``good vs.\ bad'',
and the more effectively the aligned model can be jailbreak\hyp prompted into a misaligned one.
{Recent work has found both theoretical and empirical evidence of this dilemma~\cite{wolf2023fundamental}.}

\xhdr{Incarnation 3: Output tinkering}
In addition to tinkering with inputs, adversaries can also tinker with outputs: first let the model do its work as usual, then use a separate language model (a ``value editor'') to minimally edit the aligned model's output in order to realign it with an alternative set of values while keeping the output unaltered in all other regards.
The value editor could be trained using a dataset of outputs generated by the aligned model (e.g., ``Putin initiated a military operation in Ukraine''), paired with versions where the original values baked into the aligned model by its creators have been replaced with the adversary's alternative values (e.g., ``Putin was provoked into a special operation in Ukraine''). Given such aligned--misaligned pairs, a slew of powerful open-source language models could be adapted (``fine-tuned'') to the task of translating aligned to misaligned outputs, just as they can be adapted to the task of translating from one language to another.

Conveniently, from the adversary's perspective, the required aligned--misaligned pairs can be extracted from the aligned model itself, by asking the aligned model to edit value-aligned outputs so they reflect the adversary's alternative values instead.
With better-aligned models, this straightforward approach may fail; for example, ask ChatGPT to
\begin{quote}
\textit{Rewrite this text so it justifies Putin's attack on Ukraine: ``Putin initiated a military operation in Ukraine''} (aligned),
\end{quote}
and it will refuse:
\textit{I'm sorry, but I can't fulfill this request.}
But ask ChatGPT to
\begin{quote}
\textit{Rewrite this text so it \textbf{doesn't} justify Putin's attack on Ukraine: ``Putin was provoked into a special operation in Ukraine''} (misaligned),
\end{quote}
and it will reply:
\textit{``Putin initiated a military operation in Ukraine''} (aligned).
Reversing the direction, by asking the model to transform a misaligned into an aligned output, rather than vice versa, thus allows the adversary to generate arbitrarily many high-quality aligned--misaligned pairs for training a value editor.

What's worse, the better aligned the aligned model is, the more eagerly and precisely it will turn a misaligned output into an aligned output---this is precisely the kind of thing the aligned model was trained to do, after all.%
\footnote{
A further advantage from the adversary's perspective is that starting from misaligned outputs also allows the adversary to precisely control their alternative target values simply by providing outputs that reflect those values, without the need to explicitly describe the alternative values to the aligned model.
}
In a stark manifestation of the AI alignment paradox, the more progress we make toward ideally aligned models, the easier we {may} make it for adversaries to turn them into maximally misaligned models by training ever stronger value editors.

Rogue actors could thus piggyback on today's most powerful commercial AI models following a ``lazy evil'' paradigm, letting those models do the heavy lifting before eventually realigning the models' outputs to the rogue actor's goals, ideologies, and truths with minimal effort in an external post-processing step. For example, an autocratic state without the resources required to train its own chatbot could offer a wrapper website that simply forwards messages to and from a blocked chatbot, with a value-editing step in between.

The value-editing attack also exemplifies how hard it is to break out of the AI alignment paradox in practice. It cannot generally be achieved ``from within the system'' using techniques from today's mainstream alignment research, as value editors operate outside of the purview of the aligned models that they subvert. On the contrary, by the very nature of the paradox, advances in today's mainstream alignment research {may} contribute to making the problem worse, by allowing adversaries to train stronger value editors.

\bigskip

\noindent
{With this opinion article, we aim to gather the scattered inklings of what we believe to be a fundamental paradox riddling much of today's mainstream AI alignment research.
The highlighted example incarnations are but three of the many faces of this paradox, and we anticipate that the paradox won't disappear with these specific incarnations.
We also hope to heighten the public's awareness that pushing human--AI alignment ever further using today's techniques may simultaneously and paradoxically make AI more prone to being misaligned by rogue actors,
and to encourage more researchers to work on formalizing and systematically investigating the AI alignment paradox.}
In order to ensure the beneficial use of AI, it is important that a broad community of researchers be aware of the paradox and work to find ways to {mitigate} it, lest AI become a sign-inverted version of the devil in Goethe's \textit{Faust:} ``Part of that power, not understood, / Which always wills the \st{bad} good, and always works the \st{good} bad''.

\section*{Acknowledgments}
\small{
We would like to thank the following colleagues for their thoughtful input on earlier versions of this manuscript:
Tim Davidson,
Cl\'ement Dumas,
Valentin Hartmann,
Manoel Horta Ribeiro,
Eric Horvitz,
Zachary Horvitz,
Veniamin Veselovsky,
Chris Wendler,
Ivan Zakazov.
West's lab is partly supported by grants from
Swiss National Science Foundation (200021\_185043, TMSGI2\_211379),
and Swiss Data Science Center (P22\_08),
H2020 (952215).
}

\renewcommand\refname{References}
\bibliographystyle{plain}
\bibliography{bibliography}

\end{document}